\documentclass[10pt, conference]{IEEEtran}
\IEEEoverridecommandlockouts
\usepackage{cite}
\usepackage{amsmath, amssymb, amsfonts}
\usepackage{algorithmic}
\usepackage{graphicx}
\usepackage{textcomp}
\usepackage{xcolor}
\usepackage{booktabs}
\usepackage{multirow}
\usepackage{array}
\usepackage{makecell}
\usepackage{longtable}
\usepackage{subcaption}
\usepackage{tikz}
\usepackage{pgfplots}
\pgfplotsset{compat=1.18}
\usepackage{placeins}
\usepackage{dblfloatfix}
\usepackage{url}
\usepackage{hyperref}
\usepackage{tabularx}

\def\BibTeX{{\rm B\kern-.05em{\sc i\kern-.025em b}\kern-.08em T\kern-.1667em\lower.7ex\hbox{E}\kern-.125emX}}

\begin{document}
    \title{Training Deep Stereo Matching Networks on Tree Branch Imagery: A Benchmark
    Study for Real-Time UAV Forestry Applications}
    \author{\IEEEauthorblockN{Yida Lin, Bing Xue, Mengjie Zhang} \IEEEauthorblockA{\small \textit{Centre for Data Science and Artificial Intelligence} \\ \textit{Victoria University of Wellington, Wellington, New Zealand}\\ linyida\texttt{@}myvuw.ac.nz, bing.xue\texttt{@}vuw.ac.nz, mengjie.zhang\texttt{@}vuw.ac.nz}
    \and \IEEEauthorblockN{Sam Schofield, Richard Green} \IEEEauthorblockA{\small \textit{Department of Computer Science and Software Engineering} \\ \textit{University of Canterbury, Canterbury, New Zealand}\\ sam.schofield\texttt{@}canterbury.ac.nz, richard.green\texttt{@}canterbury.ac.nz}
    }
    \maketitle
    \vspace{-1.5em}

    \begin{abstract}
        Autonomous drone-based tree pruning needs accurate, real-time depth
        estimation from stereo cameras. Depth is computed from disparity maps using
        $Z = fB/d$, so even small disparity errors cause noticeable depth
        mistakes at working distances. Building on our earlier work that identified
        DEFOM-Stereo as the best reference disparity generator for vegetation scenes,
        we present the first study to train and test ten deep stereo matching
        networks on real tree branch images. We use the Canterbury Tree Branches
        dataset---5{,}313 stereo pairs from a ZED Mini camera at 1080P and 720P---with
        DEFOM-generated disparity maps as training targets. The ten methods
        cover step-by-step refinement, 3D convolution, edge-aware attention, and
        lightweight designs. Using perceptual metrics (SSIM, LPIPS, ViTScore)
        and structural metrics (SIFT/ORB feature matching), we find that BANet-3D
        produces the best overall quality (SSIM\,=\,0.883, LPIPS\,=\,0.157), while
        RAFT-Stereo scores highest on scene-level understanding (ViTScore\,=\,0.799).
        Testing on an NVIDIA Jetson Orin Super (16\,GB, independently powered) mounted
        on our drone shows that AnyNet reaches 6.99\,FPS at 1080P---the only near-real-time
        option---while BANet-2D gives the best quality--speed balance at 1.21\,FPS.
        We also compare 720P and 1080P processing times to guide resolution choices
        for forestry drone systems.
    \end{abstract}

    \vspace{-1em}
    \section{Introduction}

    Radiata pine (\textit{Pinus radiata}) is the main plantation tree in New Zealand,
    where forestry adds NZ\$3.6 billion (1.3\%) to the national economy. Trees
    need regular pruning to produce high-quality wood, but pruning by hand is dangerous
    due to fall risks and chainsaw injuries. Drones that prune automatically~\cite{lin2024branch,steininger2025timbervision}
    offer a safer option, yet they need centimeter-level depth accuracy to position
    cutting tools at 1--2\,m distances. In a stereo camera system, depth is not
    measured directly. Instead, it comes from \textit{disparity maps}---images that
    record how far each pixel shifts horizontally between the left and right camera
    views. Forest scenes are much harder than city or indoor environments
    because they contain thin overlapping branches, repeating textures, sharp
    depth changes, and large lighting differences. These challenges require
    training on vegetation-specific data rather than using general-purpose
    pretrained models.

    \textbf{Disparity Maps and Depth Recovery.} A disparity map $D \in \mathbb{R}
    ^{H \times W}$ stores, for each pixel $(x,y)$ in the left image, how far its
    matching pixel $(x - D(x,y), y)$ is shifted in the right image. Depth $Z$ is
    then calculated by triangulation:
    \begin{equation}
        Z(x,y) = \frac{f \cdot B}{D(x,y)}\label{eq:depth}
    \end{equation}
    where $f$ is the focal length and $B$ is the distance between the two cameras
    (the baseline). Since depth is inversely proportional to disparity, even small
    disparity errors lead to increasingly large depth errors at longer distances.
    With our ZED Mini camera ($B = 63$\,mm), a 1-pixel disparity mistake at 1.5\,m
    branch distance produces roughly 2.3\,cm of depth error at 1080P resolution.
    This highlights why accurate disparity prediction is the main goal of this
    work. Once disparity is known, exact depth follows directly from Eq.~\eqref{eq:depth}
    and the camera's calibration settings.

    In our earlier benchmark study~\cite{lin2026benchmark}, we found that DEFOM-Stereo~\cite{jiang2025defom}
    is the best method for producing reference disparity maps (pseudo-ground-truth)
    on vegetation scenes. We tested it without any task-specific training across
    several standard datasets, and it ranked first on average (rank 1.75 across
    ETH3D~\cite{schops2017eth3d}, KITTI~\cite{geiger2012kitti}, and Middlebury~\cite{scharstein2014middlebury})
    with no major failures. This consistency makes DEFOM well-suited for
    creating training labels. The current paper uses those DEFOM-generated disparity
    maps as training targets to build and evaluate stereo matching models designed
    for vegetation scenes.

    Deep stereo methods perform well on standard benchmarks when trained on synthetic
    data like Scene Flow~\cite{mayer2016large}, but their accuracy drops sharply
    on forestry images. Our zero-shot generalization study~\cite{lin2025generalization}
    confirmed this gap: methods trained only on synthetic data show large
    performance differences when applied to vegetation scenes. This happens because
    synthetic city scenes look very different from real vegetation. Training on
    forest-specific data would fix this, but collecting accurate disparity maps with
    LiDAR in forest canopies is not practical---branches cause heavy blocking and
    their thin, complex shapes are hard to scan. Our approach bypasses this problem:
    we use DEFOM's high-quality predictions as reference labels, allowing large-scale
    training on real forestry data without costly LiDAR equipment.

    Three main questions drive this study. \textit{First}, which network design
    produces the best disparity maps---as measured by perceptual and structural
    metrics---after training on vegetation data? \textit{Second}, which design offers
    the best balance between quality and processing speed for drone deployment on
    low-power hardware? \textit{Third}, how does image resolution (1080P vs.\ 720P)
    affect processing speed and real-world usability on drones with limited computing
    resources?

    To answer these questions, we train and test ten deep stereo methods on the
    Canterbury Tree Branches dataset: RAFT-Stereo~\cite{lipson2021raft} (step-by-step
    refinement), PSMNet~\cite{chang2018psmnet} and GwcNet~\cite{guo2019gwcnet} (3D
    convolution networks), MoCha-Stereo~\cite{chen2024mocha} (motion and channel
    attention), BANet-2D and BANet-3D~\cite{tankovich2021hitnet} (edge-aware attention
    with 2D/3D cost processing), IGEV-RT~\cite{xu2023igev} (fast geometry-based
    refinement), DeepPruner~\cite{duggal2019deeppruner} (search space reduction),
    DCVSMNet~\cite{tahmasebi2025dcvsmnet} (dual cost volume), and AnyNet~\cite{wang2019anynet}
    (multi-stage prediction). All models are trained with DEFOM reference labels
    and tested on held-out scenes using perceptual and structural metrics. We
    also run all models on an NVIDIA Jetson Orin Super mounted on our test drone
    to measure real-world processing speed at both 1080P and 720P. Our contributions
    are:

    \begin{itemize}
        \item \textbf{First vegetation-focused stereo benchmark}: We create the Canterbury
            Tree Branches dataset with DEFOM reference labels as a training and testing
            resource for forestry use, removing the need for costly LiDAR data
            collection.

        \item \textbf{Broad ten-method comparison}: We train and test ten stereo
            methods from six design families using perceptual (SSIM, LPIPS,
            ViTScore) and structural (SIFT/ORB feature matching) metrics. BANet-3D
            emerges as the top-quality method.

        \item \textbf{Quality--speed trade-off analysis}: We map out the best trade-off
            boundary between disparity quality and processing speed, finding
            that only AnyNet, BANet-2D, and BANet-3D offer unbeatable
            combinations.

        \item \textbf{Real-world drone deployment}: We show practical results on
            an independently-powered NVIDIA Jetson Orin Super (16\,GB) with live
            ZED Mini stereo input at 1080P and 720P, providing speed profiles
            and deployment guidelines for autonomous pruning.
    \end{itemize}
    \vspace{-1em}
    \section{Related Work}

    \subsection{Deep Stereo Matching Architectures}

    Deep learning has changed how stereo matching works by enabling end-to-end trainable
    systems. Early methods used 3D convolution networks: DispNet~\cite{mayer2016large}
    showed that networks can directly predict disparity, GC-Net~\cite{kendall2017gcnet}
    added 3D convolutions to process matching costs, and PSMNet~\cite{chang2018psmnet}
    used multi-scale pooling to capture context at different levels. GwcNet~\cite{guo2019gwcnet}
    introduced group-wise correlation to build matching costs more efficiently.

    Step-by-step refinement methods, inspired by motion estimation, handle large
    disparity ranges through repeated updates. RAFT-Stereo~\cite{lipson2021raft}
    uses recurrent updates with multi-scale matching and reaches top accuracy,
    though at high computing cost. IGEV-Stereo~\cite{xu2023igev} combines repeated
    updates with geometry-based features; its faster version, IGEV-RT, reduces
    the number of update steps to speed things up while keeping reasonable
    quality.

    Edge-aware attention methods add boundary-sensitive filtering to the
    matching process. BANet~\cite{tankovich2021hitnet} comes in both 2D and 3D versions,
    performing well by respecting depth edges through efficient bilateral
    operations.

    Lightweight models focus on running fast on limited hardware. AnyNet~\cite{wang2019anynet}
    predicts disparity in stages from coarse to fine. DeepPruner~\cite{duggal2019deeppruner}
    narrows down the search area to save computation. DCVSMNet~\cite{tahmasebi2025dcvsmnet}
    uses two separate cost volumes for matching. MoCha-Stereo~\cite{chen2024mocha}
    applies motion and channel attention to improve matching accuracy. These designs
    are especially relevant for drones, where computing power is tightly limited.

    \subsection{Pseudo-Ground-Truth for Stereo Training}

    Training stereo networks requires dense disparity labels for every pixel.
    These are usually obtained with LiDAR scanners~\cite{geiger2012kitti} or
    structured light systems~\cite{scharstein2014middlebury}. However, such
    equipment cannot work well in many real settings---especially forest
    canopies, where branches block the scanner and prevent accurate measurements.

    An alternative is to use high-quality predictions from strong existing models
    as training labels (pseudo-ground-truth). Large-scale models like DEFOM-Stereo~\cite{jiang2025defom}
    and Depth Anything~\cite{yang2024depth} generalize well across different
    scenes, making them good candidates for label generation. In our earlier
    work~\cite{lin2026benchmark,lin2025generalization}, we compared several stereo
    methods and confirmed that DEFOM gives the most consistent results on vegetation
    scenes.

    \subsection{Stereo Vision for UAV Forestry}

    Drone-based stereo vision has been used for navigation~\cite{fraundorfer2012vision},
    3D mapping~\cite{nex2014uav}, and obstacle avoidance~\cite{barry2015pushbroom}.
    Recent studies on branch detection~\cite{lin2024branch,lin2025yolosgbm},
    segmentation~\cite{lin2025segmentation}, and disparity parameter optimization
    using genetic algorithms~\cite{lin2025ga} show that deep learning and stereo
    cameras have clear potential for forestry tasks. Still, most existing systems
    use traditional matching algorithms or pretrained models without adapting them
    to vegetation. This work fills that gap by training modern stereo networks
    directly on tree imagery.

    \subsection{Real-Time Stereo on Embedded Platforms}

    Running stereo networks on small embedded computers means balancing accuracy
    against speed. NVIDIA Jetson boards have become the standard choice for
    drone perception~\cite{liu2017survey}, and they support TensorRT for faster processing.
    Earlier work showed that real-time stereo is possible at lower resolutions~\cite{wang2019anynet},
    but getting both high resolution and real-time speed remains hard. In this
    work, we systematically test all ten network designs at both 720P and 1080P on
    Jetson hardware with its own power supply, matching the conditions of actual
    drone flights.

    \section{Methodology}

    \subsection{Problem Formulation}

    Stereo matching estimates the per-pixel disparity from a pair of aligned
    stereo images. Given left and right images $I_{L}, I_{R}\in \mathbb{R}^{H
    \times W \times 3}$, the task is to produce a disparity map
    $D \in \mathbb{R}^{H \times W}$ where each pixel $(x,y)$ in the left image
    matches pixel $(x - D(x,y), y)$ in the right image.

    Disparity and depth follow the triangulation formula (Eq.~\eqref{eq:depth}).
    Because depth $Z$ is inversely related to disparity $D$, objects close to
    the camera (like branches at 1--2\,m) produce large disparity values and fine
    depth detail, while distant background gives small disparity with limited
    depth precision. For our ZED Mini stereo camera with baseline $B = 63$\,mm,
    operating at 1080P ($f \approx 700$\,px) and 720P ($f \approx 350$\,px), this
    relationship provides sub-centimeter depth resolution at the 1--2\,m range
    critical for pruning. We therefore train networks $f_{\theta}$ to predict disparity,
    and then convert to metric depth using Eq.~\eqref{eq:depth}.

    \subsection{Dataset}

    \textbf{Canterbury Tree Branches Dataset}: We use 5{,}313 stereo pairs taken
    with a ZED Mini camera (63\,mm baseline) on a drone in Canterbury, New
    Zealand (March--October 2024). The camera captures synchronized left and
    right images at 1920$\times$1080 (1080P) and 1280$\times$720 (720P). We train
    and evaluate quality at 1080P, and measure processing speed at both
    resolutions. The images were carefully chosen from hundreds of thousands of raw
    captures by filtering for motion blur, exposure quality, alignment accuracy,
    and scene variety.

    \textbf{Pseudo-Ground-Truth Generation}: Following our earlier study~\cite{lin2026benchmark},
    we run DEFOM-Stereo~\cite{jiang2025defom} on all 5{,}313 pairs to produce reference
    disparity maps. DEFOM was chosen for its consistent performance across
    datasets (average rank 1.75 across ETH3D, KITTI, and Middlebury) and its
    strong results on vegetation scenes. These outputs serve as the training targets
    for all ten methods.

    \textbf{Train/Validation/Test Split}: We divide the data into training (4{,}250
    pairs, 80\%), validation (531 pairs, 10\%), and test (532 pairs, 10\%) sets.
    The splits cover different times and locations to prevent the models from
    memorizing specific scenes or lighting conditions.

    \subsection{Evaluated Methods}

    We pick ten methods from six design families for training and testing (Table~\ref{tab:methods}).

    \begin{table}[htbp]
        \caption{Evaluated Stereo Matching Methods}
        \label{tab:methods}
        \centering
        \small
        \begin{tabular}{lcc}
            \toprule \textbf{Method}                   & \textbf{Type}   & \textbf{Key Feature}     \\
            \midrule RAFT-Stereo~\cite{lipson2021raft} & Iterative       & Recurrent refinement     \\
            PSMNet~\cite{chang2018psmnet}              & 3D CNN          & Spatial pyramid pooling  \\
            GwcNet~\cite{guo2019gwcnet}                & 3D CNN          & Group-wise correlation   \\
            MoCha-Stereo~\cite{chen2024mocha}          & Attention       & Motion-channel attention \\
            BANet-2D~\cite{tankovich2021hitnet}        & Bilateral Attn. & 2D cost filtering        \\
            BANet-3D~\cite{tankovich2021hitnet}        & Bilateral Attn. & 3D cost filtering        \\
            IGEV-RT~\cite{xu2023igev}                  & Iterative (RT)  & Geometry encoding volume \\
            DeepPruner~\cite{duggal2019deeppruner}     & PatchMatch      & Differentiable pruning   \\
            DCVSMNet~\cite{tahmasebi2025dcvsmnet}      & Dual CV         & Dual cost volume         \\
            AnyNet~\cite{wang2019anynet}               & Hierarchical    & Anytime prediction       \\
            \bottomrule
        \end{tabular}
    \end{table}

    \subsection{Training Protocol}

    All models start from weights pretrained on Scene Flow~\cite{mayer2016large}
    and are then fine-tuned on our Tree Branches training set. We keep the same
    settings across methods where possible:

    \begin{itemize}
        \item \textbf{Optimizer}: AdamW with weight decay $10^{-4}$

        \item \textbf{Learning rate}: $10^{-4}$ with cosine annealing

        \item \textbf{Batch size}: 4 (adjusted per GPU memory)

        \item \textbf{Training epochs}: 100 with early stopping (patience 10)

        \item \textbf{Input resolution}: 512$\times$256 crops during training

        \item \textbf{Data augmentation}: Random crops, color jitter, horizontal
            flip
    \end{itemize}

    \textbf{Loss Function}: We train with smooth L1 loss, which measures the
    difference between predicted and reference disparity values:
    \begin{equation}
        \mathcal{L}= \frac{1}{N}\sum_{i=1}^{N}\text{smooth}_{L_1}(d_{i}- \hat{d}_{i}
        )
    \end{equation}
    where $\hat{d}_{i}$ is the DEFOM reference value. For methods that produce
    outputs at multiple scales (PSMNet, GwcNet, BANet-3D), we add weighted losses
    at each scale.

    \subsection{Evaluation Metrics}

    Instead of only checking pixel-by-pixel error, we use several complementary metrics
    that capture both visual quality and structural accuracy of the predicted disparity
    maps compared to the DEFOM reference.

    \textbf{Structural Similarity (SSIM)}~\cite{wang2004ssim} checks how well
    the brightness, contrast, and structure of predicted maps match the
    reference. Higher SSIM means better preservation of disparity gradients and boundaries
    between regions.

    \textbf{Learned Perceptual Image Patch Similarity (LPIPS)}~\cite{zhang2018lpips}
    uses deep network features to measure how different two images look to a
    human. Lower LPIPS means the prediction looks more like the reference,
    catching fine details that simple pixel comparisons miss.

    \textbf{ViTScore} measures high-level similarity using Vision Transformer features.
    Higher ViTScore means the predicted map keeps the overall geometric layout---branch
    shapes, depth layers, and blocked regions---intact.

    \textbf{Feature Matching Ratios}: We detect SIFT and ORB keypoints in both
    predicted and reference maps, then count how many match successfully (using
    the Lowe ratio test). Higher match ratios mean the prediction faithfully
    reproduces structural features like edges, corners, and texture patterns. This
    matters for later tasks such as branch detection and distance measurement.

    \textbf{Inference Latency}: We record average frames per second (FPS) and per-frame
    delay on the target embedded board, which is essential for real-time drone
    use.

    \subsection{Hardware Platforms}

    \textbf{Training}: NVIDIA Quadro RTX 6000 GPU (24\,GB VRAM) with PyTorch 2.6.0.

    \textbf{On-Drone Deployment}: NVIDIA Jetson Orin Super (16\,GB shared memory),
    mounted on our test drone and powered by its own dedicated battery, separate
    from the flight battery. This separate power design prevents the computing load
    from draining the flight reserves, avoiding performance drops during long
    flights. The Jetson takes live stereo input from the ZED Mini camera at both
    1080P (1920$\times$1080) and 720P (1280$\times$720).

    \section{Experimental Results}

    \subsection{Comprehensive Quality and Speed Evaluation}

    Table~\ref{tab:combined_results} shows the quality and speed results for all
    ten methods trained on the Canterbury Tree Branches dataset with DEFOM reference
    labels. Quality scores come from the held-out test set (532 pairs, 1080P),
    and speed numbers are from the NVIDIA Jetson Orin Super at 1080P.

    \begin{table*}
        [htbp]
        \caption{Comprehensive Evaluation: Disparity Quality Metrics and Inference
        Speed on NVIDIA Jetson Orin Super (1080P)}
        \label{tab:combined_results}
        \centering
        \small
        \begin{tabular}{lccccccc}
            \toprule \textbf{Method} & \textbf{SSIM}$\uparrow$ & \textbf{LPIPS}$\downarrow$ & \textbf{ViTScore}$\uparrow$ & \textbf{SIFT Ratio}$\uparrow$ & \textbf{ORB Ratio}$\uparrow$ & \textbf{FPS}$\uparrow$ & \textbf{Latency (ms)}$\downarrow$ \\
            \midrule BANet-3D        & \textbf{0.883}          & \textbf{0.157}             & 0.790                       & \textbf{0.274}                & \textbf{0.162}               & 0.71                   & 1408                              \\
            MoCha-Stereo             & 0.848                   & 0.221                      & 0.701                       & 0.111                         & 0.044                        & 0.14                   & 7143                              \\
            BANet-2D                 & 0.816                   & 0.245                      & 0.724                       & 0.171                         & 0.072                        & 1.21                   & 826                               \\
            GwcNet                   & 0.811                   & 0.250                      & 0.750                       & 0.157                         & 0.047                        & 0.23                   & 4348                              \\
            PSMNet                   & 0.810                   & 0.212                      & 0.786                       & 0.181                         & 0.080                        & 0.11                   & 9091                              \\
            IGEV-RT                  & 0.794                   & 0.297                      & 0.558                       & 0.071                         & 0.020                        & 0.24                   & 4167                              \\
            DCVSMNet                 & 0.769                   & 0.347                      & 0.509                       & 0.047                         & 0.012                        & 0.36                   & 2778                              \\
            AnyNet                   & 0.766                   & 0.434                      & 0.196                       & 0.065                         & 0.012                        & \textbf{6.99}          & \textbf{143}                      \\
            RAFT-Stereo              & 0.763                   & 0.235                      & \textbf{0.799}              & 0.245                         & 0.108                        & 0.07                   & 14286                             \\
            DeepPruner               & 0.720                   & 0.542                      & 0.397                       & 0.039                         & 0.011                        & 0.34                   & 2941                              \\
            \bottomrule
        \end{tabular}
        \vspace{0.5em}
        \begin{flushleft}
            \footnotesize{All methods trained on Tree Branches training set with DEFOM pseudo-ground-truth. Quality metrics evaluated on test set (532 pairs, 1080P). FPS measured on NVIDIA Jetson Orin Super at 1080P. Bold indicates best per column. SIFT/ORB Ratio = proportion of successfully matched keypoints (Lowe ratio test). $\uparrow$: higher is better; $\downarrow$: lower is better.}
        \end{flushleft}
    \end{table*}

    \subsubsection{Quality Analysis}

    \textbf{Edge-Aware Attention Methods}: BANet-3D delivers the best disparity
    quality on four out of five metrics: highest SSIM (0.883), lowest LPIPS (0.157),
    and highest SIFT (0.274) and ORB (0.162) match ratios. Its 3D cost filtering
    keeps depth edges sharp and preserves thin branch details, producing maps that
    closely match the DEFOM reference both visually and structurally. BANet-2D
    is a lighter version with solid quality (SSIM\,=\,0.816, LPIPS\,=\,0.245) and
    runs 1.7$\times$ faster.

    \textbf{Step-by-Step Refinement Methods}: RAFT-Stereo scores highest on
    ViTScore (0.799) and second on SIFT matching (0.245), showing that repeated updates
    capture the overall depth layout and scene structure well. However, its SSIM
    (0.763) is near the bottom, meaning that while the big-picture depth is good,
    pixel-level smoothness suffers. IGEV-RT, designed for faster processing,
    reaches moderate quality (SSIM\,=\,0.794, ViTScore\,=\,0.558) but still only
    manages 0.24\,FPS at 1080P---far from real-time.

    \textbf{3D Convolution Methods}: PSMNet gets the second-best LPIPS (0.212) and
    third-best ViTScore (0.786), showing that its multi-scale pooling captures
    context well for vegetation. GwcNet performs similarly (SSIM\,=\,0.811, ViTScore\,=\,0.750)
    and runs slightly faster.

    \textbf{Attention-Based}: MoCha-Stereo reaches the second-highest SSIM (0.848),
    showing that its motion and channel attention handles the repeating textures
    and fine details common in tree scenes.

    \textbf{Lightweight Methods}: AnyNet is the fastest (6.99\,FPS, 143\,ms delay)
    but at a clear quality cost---it has the worst ViTScore (0.196) and highest
    LPIPS (0.434). DeepPruner shows the weakest quality overall (SSIM\,=\,0.720,
    LPIPS\,=\,0.542), suggesting that its search space reduction approach may throw
    away important matches in complex vegetation.

    \subsubsection{Inference Speed Analysis}

    Only AnyNet (6.99\,FPS) comes close to real-time at 1080P on the Jetson Orin
    Super. BANet-2D reaches 1.21\,FPS, which is enough for tasks that can
    tolerate some delay, such as approach planning or stationary branch assessment.
    All other methods run below 1\,FPS at full resolution, so they would need either
    lower resolution or offline processing to be practical.

    \subsection{Resolution-Dependent Latency Comparison}

\begin{figure}[htbp]
        \centering
        \includegraphics[width=\columnwidth]{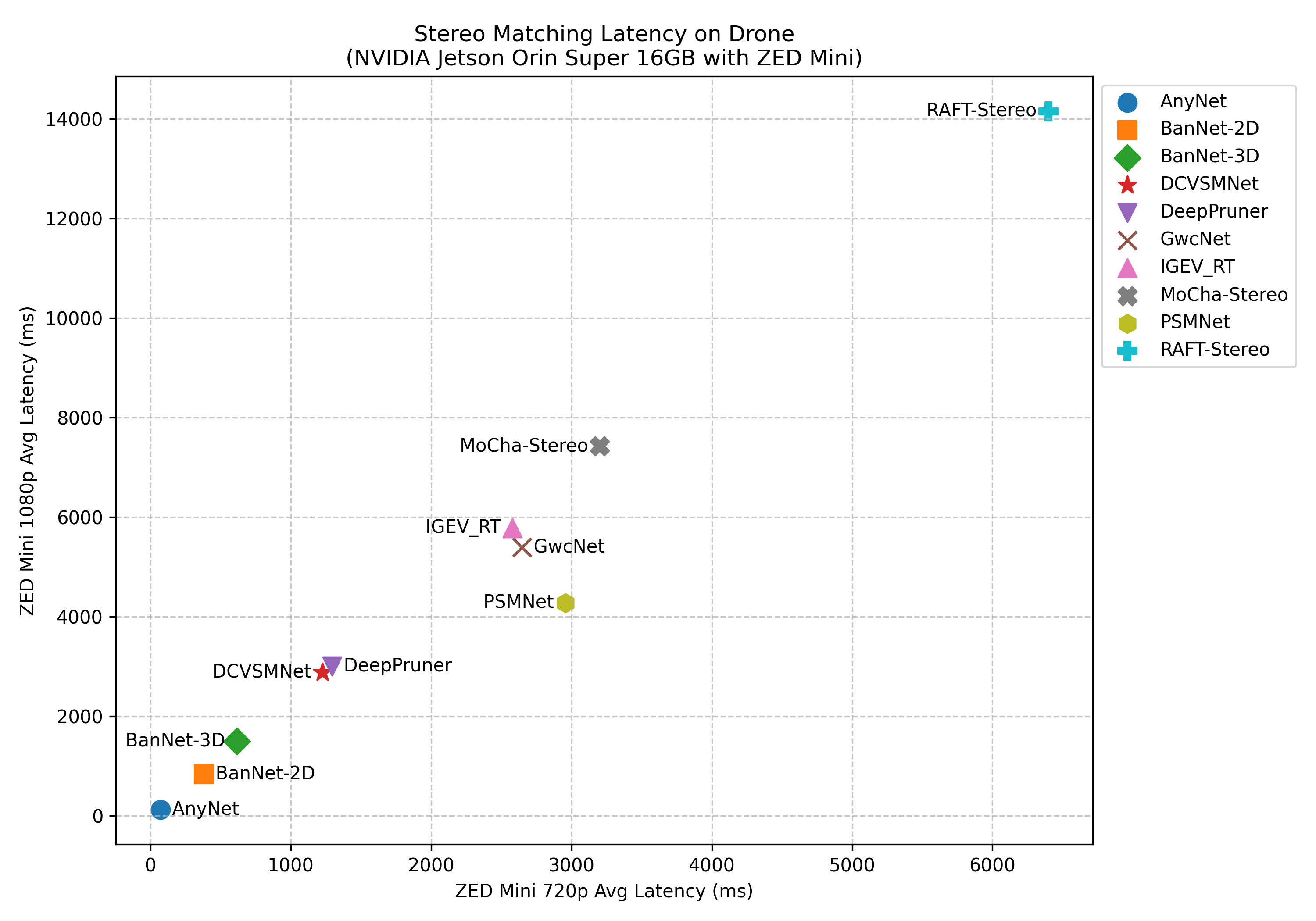}
        \caption{Inference latency comparison between 720P and 1080P modes on
        NVIDIA Jetson Orin Super. Live stereo input from ZED Mini camera. Lower
        latency is better; dashed line indicates 10\,FPS real-time threshold.}
        \label{fig:latency_comparison}
    \end{figure}

    Fig.~\ref{fig:latency_comparison} compares processing times at 720P (1280$\times$720)
    and 1080P (1920$\times$1080) on the Jetson Orin Super, using live video from
    the ZED Mini camera during drone flights.

    Dropping from 1080P to 720P cuts the pixel count by 56\%, which gives noticeable
    speed gains for most methods. AnyNet benefits the most, getting much closer to
    usable real-time speeds at 720P. Heavy methods like RAFT-Stereo and PSMNet
    stay too slow even at the lower resolution, showing that network design matters
    more than resolution for on-drone speed.

    \subsection{Quality--Speed Trade-off and Pareto Analysis}

\begin{figure}[htbp]
        \centering
        \begin{tikzpicture}
            \begin{axis}[
                xlabel={FPS on Jetson Orin Super (1080P) $\rightarrow$},
                ylabel={SSIM $\rightarrow$},
                xmin=0,
                xmax=8.5,
                ymin=0.70,
                ymax=0.92,
                grid=both,
                minor grid style={gray!15},
                major grid style={gray!30},
                width=\columnwidth,
                height=0.75\columnwidth,
                every axis label/.style={font=\small},
                tick label style={font=\footnotesize},
            ]
                \addplot[red!60, thick, dashed, no markers] coordinates { (0.71, 0.883) (1.21, 0.816) (6.99, 0.766) };
                \addplot[only marks, mark=star, mark size=4pt, red!80!black,
                fill=red!30] coordinates { (0.71, 0.883) (1.21, 0.816) (6.99, 0.766) };
                \addplot[only marks, mark=*, mark size=2.5pt, blue!80!black, fill=blue!40]
                coordinates
                { (0.07, 0.763) (0.11, 0.810) (0.14, 0.848) (0.23, 0.811) (0.24, 0.794) (0.34, 0.720) (0.36, 0.769) };
                \node[anchor=south, font=\scriptsize\bfseries, red!80!black]
                    at
                    (axis cs:0.71,0.887)
                    {BANet-3D};
                \node[
                    anchor=south west,
                    font=\scriptsize\bfseries,
                    red!80!black
                ] at (axis cs:1.28,0.816) {BANet-2D};
                \node[anchor=south, font=\scriptsize\bfseries, red!80!black]
                    at
                    (axis cs:6.99,0.772)
                    {AnyNet};
                \node[anchor=north, font=\scriptsize, blue!70!black]
                    at
                    (axis cs:0.07,0.759)
                    {RAFT-Stereo};
                \node[anchor=south east, font=\scriptsize, blue!70!black]
                    at
                    (axis cs:0.08,0.812)
                    {PSMNet};
                \node[anchor=south, font=\scriptsize, blue!70!black]
                    at
                    (axis cs:0.14,0.854)
                    {MoCha-Stereo};
                \node[anchor=west, font=\scriptsize, blue!70!black]
                    at
                    (axis cs:0.30,0.811)
                    {GwcNet};
                \node[anchor=north west, font=\scriptsize, blue!70!black]
                    at
                    (axis cs:0.26,0.790)
                    {IGEV-RT};
                \node[anchor=north, font=\scriptsize, blue!70!black]
                    at
                    (axis cs:0.34,0.714)
                    {DeepPruner};
                \node[anchor=west, font=\scriptsize, blue!70!black]
                    at
                    (axis cs:0.43,0.769)
                    {DCVSMNet};
                \addlegendimage{star, mark size=4pt, red!80!black, fill=red!30}
                \addlegendentry{Pareto-optimal} \addlegendimage{mark=*, mark size=2.5pt, blue!80!black, fill=blue!40}
                \addlegendentry{Dominated} \legend{}
            \end{axis}
            \node[
                anchor=south east,
                draw=gray!50,
                fill=white,
                rounded corners=2pt,
                inner sep=3pt,
                font=\scriptsize
            ]
                at
                (6.8,0.3)
                { \begin{tabular}{@{}cl@{}}{\color{red!80!black}$\bigstar$} & Pareto-optimal \\ {\color{blue!80!black}$\bullet$} & Dominated\end{tabular} };
        \end{tikzpicture}
        \caption{Quality--speed trade-off (SSIM vs.\ FPS) for all ten methods on
        the Jetson Orin Super at 1080P. Red stars and dashed line mark the Pareto
        frontier: BANet-3D (best quality), BANet-2D (balanced), AnyNet (fastest).
        Blue circles indicate dominated methods.}
        \label{fig:pareto}
    \end{figure}
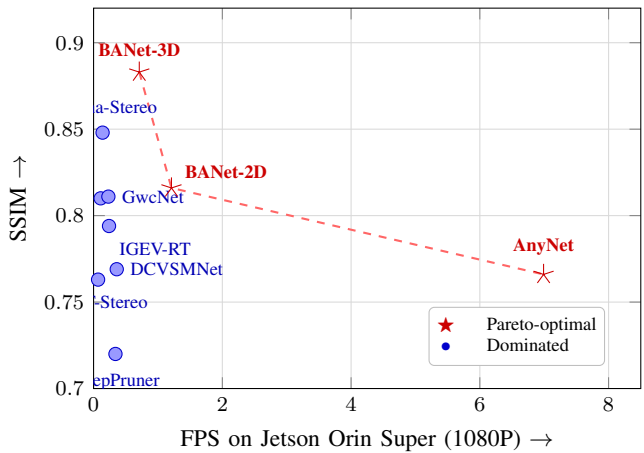

    Fig.~\ref{fig:pareto} plots quality (SSIM) against speed (FPS) for all ten methods,
    showing which ones offer the best combinations.

    \textbf{Key Finding}: Only three methods sit on the best trade-off line:
    BANet-3D (highest quality, 0.71\,FPS), BANet-2D (balanced, 1.21\,FPS), and
    AnyNet (fastest, 6.99\,FPS). Every other method is outperformed---it is
    either slower and lower-quality than one of these three, or offers no unique
    advantage. This gives clear guidance for deployment:

    \begin{itemize}
        \item \textit{When quality matters most} (offline mapping, detailed
            inspection): use BANet-3D.

        \item \textit{When balance is needed} (approach planning, slow-speed
            manoeuvres): use BANet-2D.

        \item \textit{When speed is critical} (closed-loop control, obstacle
            avoidance): use AnyNet, accepting lower disparity quality.
    \end{itemize}

    \subsection{Qualitative Results}

\begin{figure*}[htbp]
        \centering
        \begin{subfigure}
            [b]{0.16\textwidth}
            \centering
            \includegraphics[width=\textwidth]{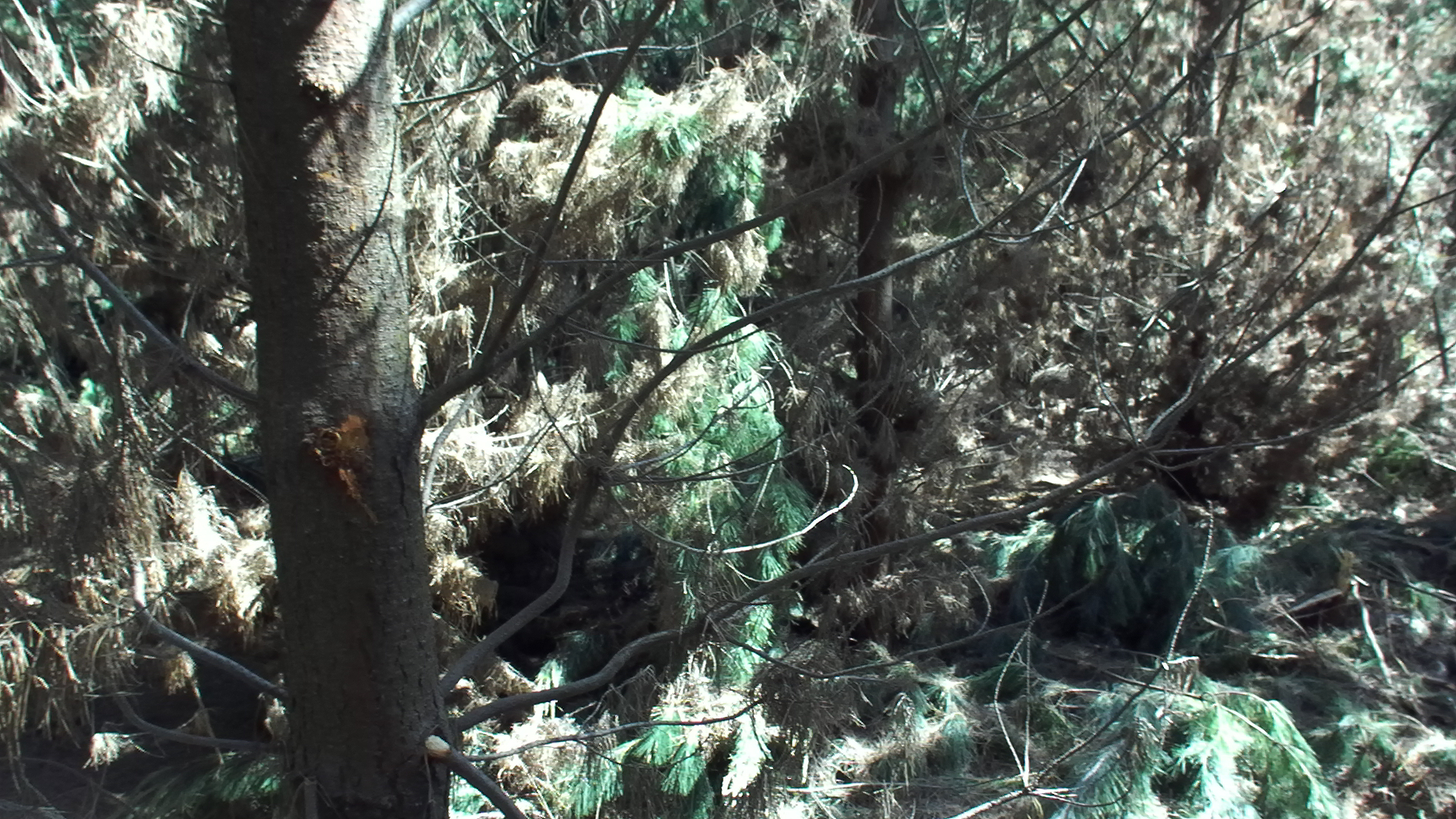}
            \caption{Left Image}
        \end{subfigure}
        \begin{subfigure}
            [b]{0.16\textwidth}
            \centering
            \includegraphics[width=\textwidth]{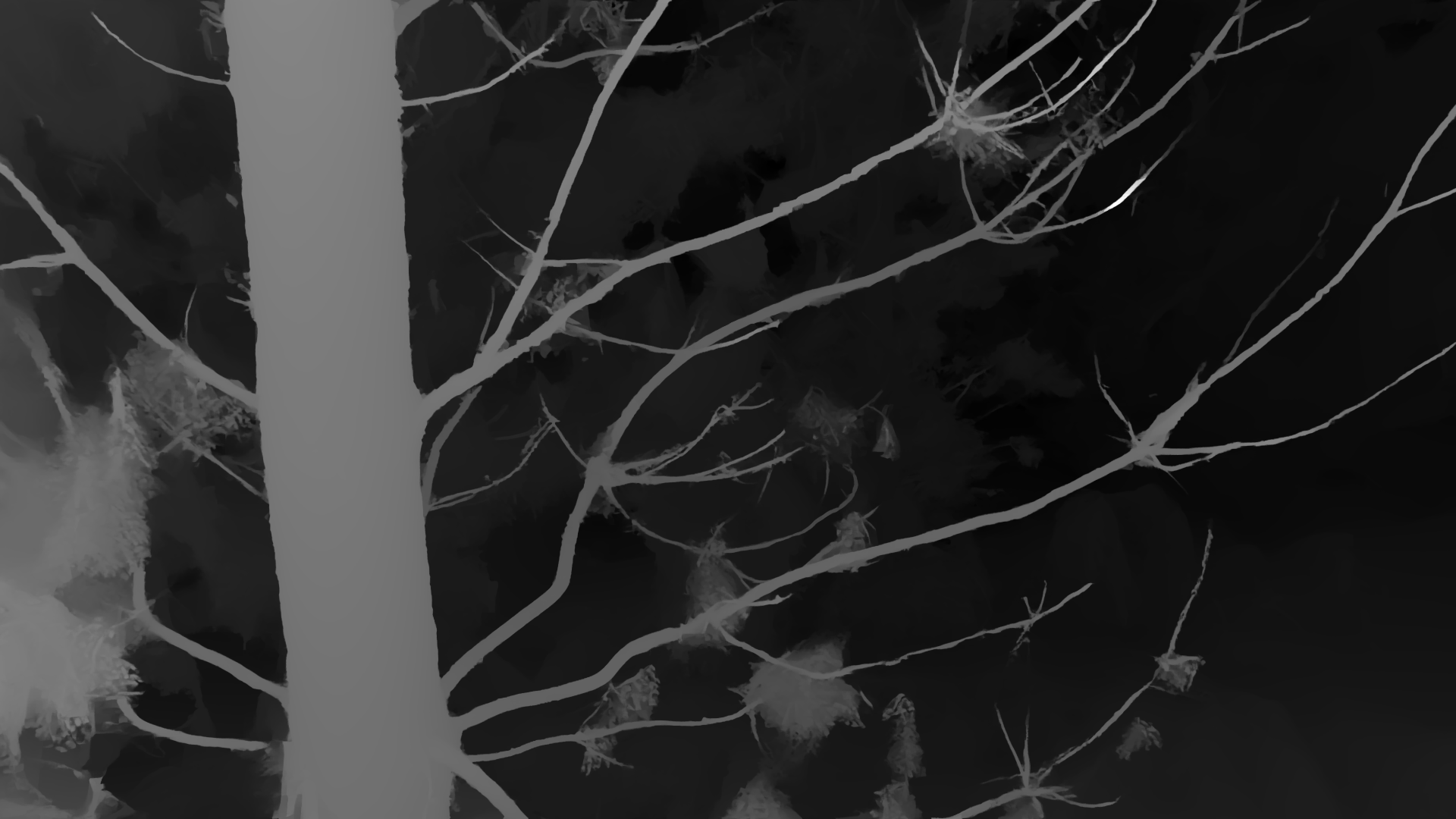}
            \caption{Ground Truth}
        \end{subfigure}
        \begin{subfigure}
            [b]{0.16\textwidth}
            \centering
            \includegraphics[width=\textwidth]{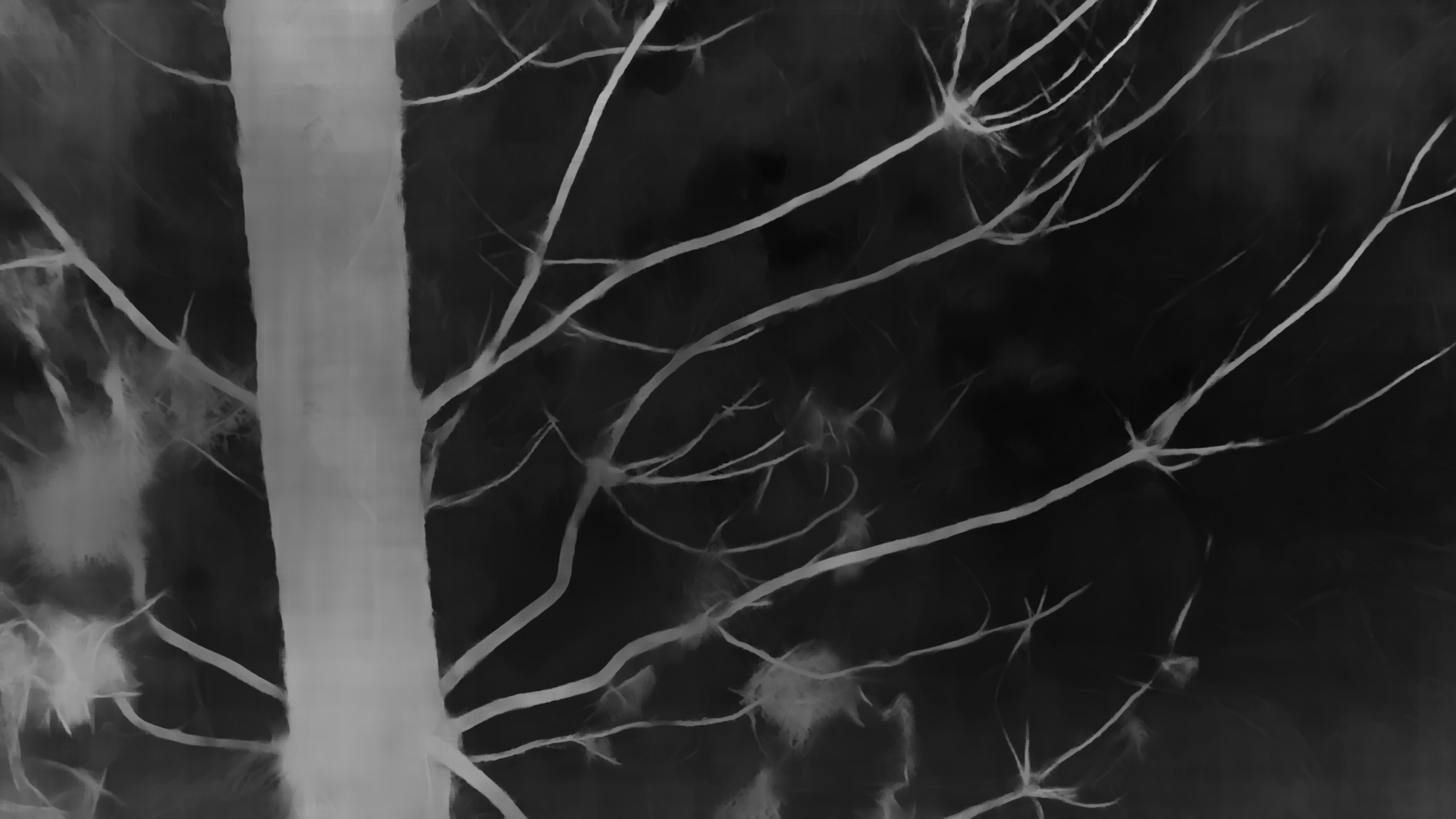}
            \caption{BANet-3D}
        \end{subfigure}
        \begin{subfigure}
            [b]{0.16\textwidth}
            \centering
            \includegraphics[width=\textwidth]{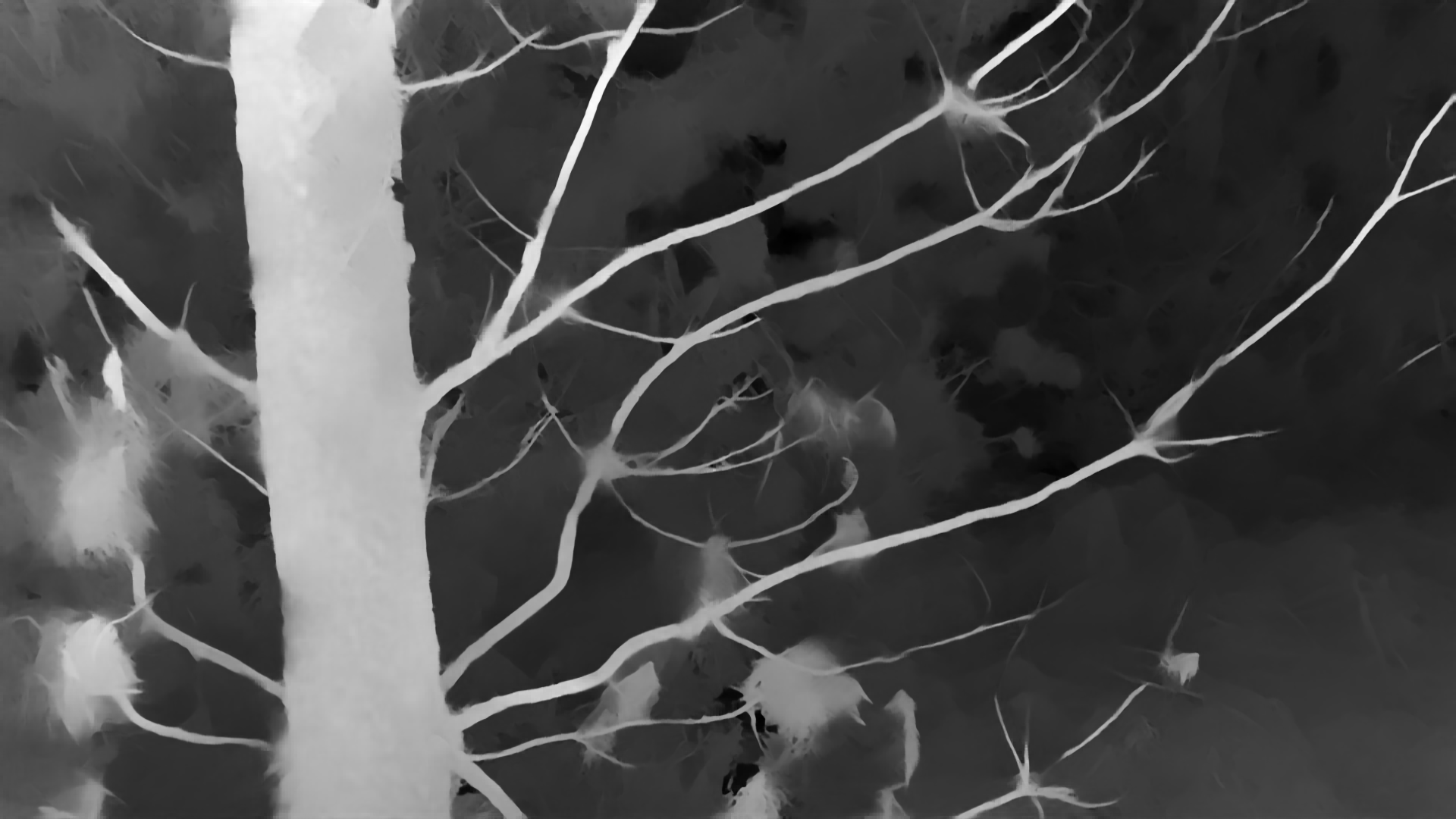}
            \caption{RAFT-Stereo}
        \end{subfigure}
        \begin{subfigure}
            [b]{0.16\textwidth}
            \centering
            \includegraphics[width=\textwidth]{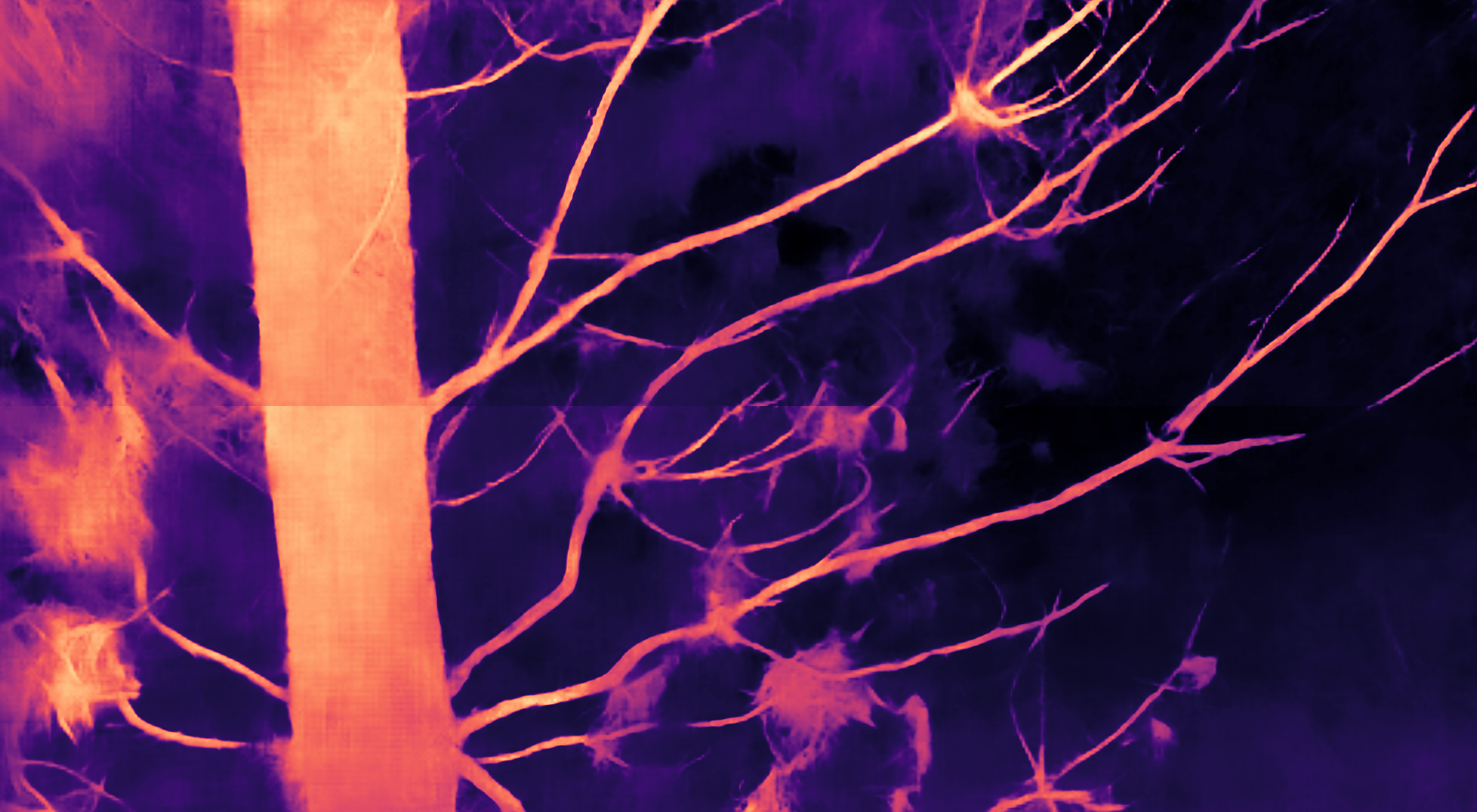}
            \caption{PSMNet}
        \end{subfigure}
        \begin{subfigure}
            [b]{0.16\textwidth}
            \centering
            \includegraphics[width=\textwidth]{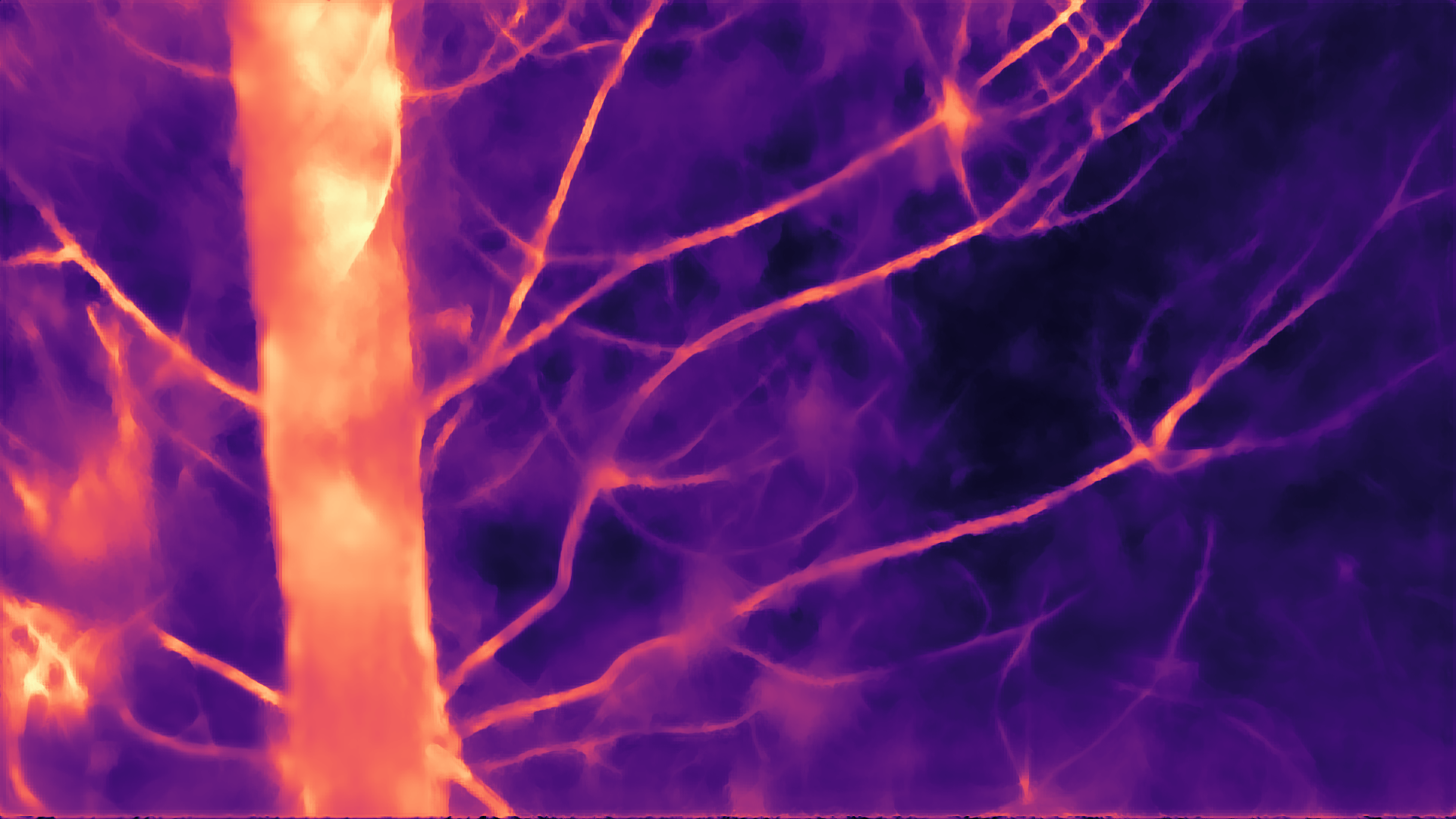}
            \caption{MoCha}
        \end{subfigure}
        \\[0.5em]
        \begin{subfigure}
            [b]{0.16\textwidth}
            \centering
            \includegraphics[width=\textwidth]{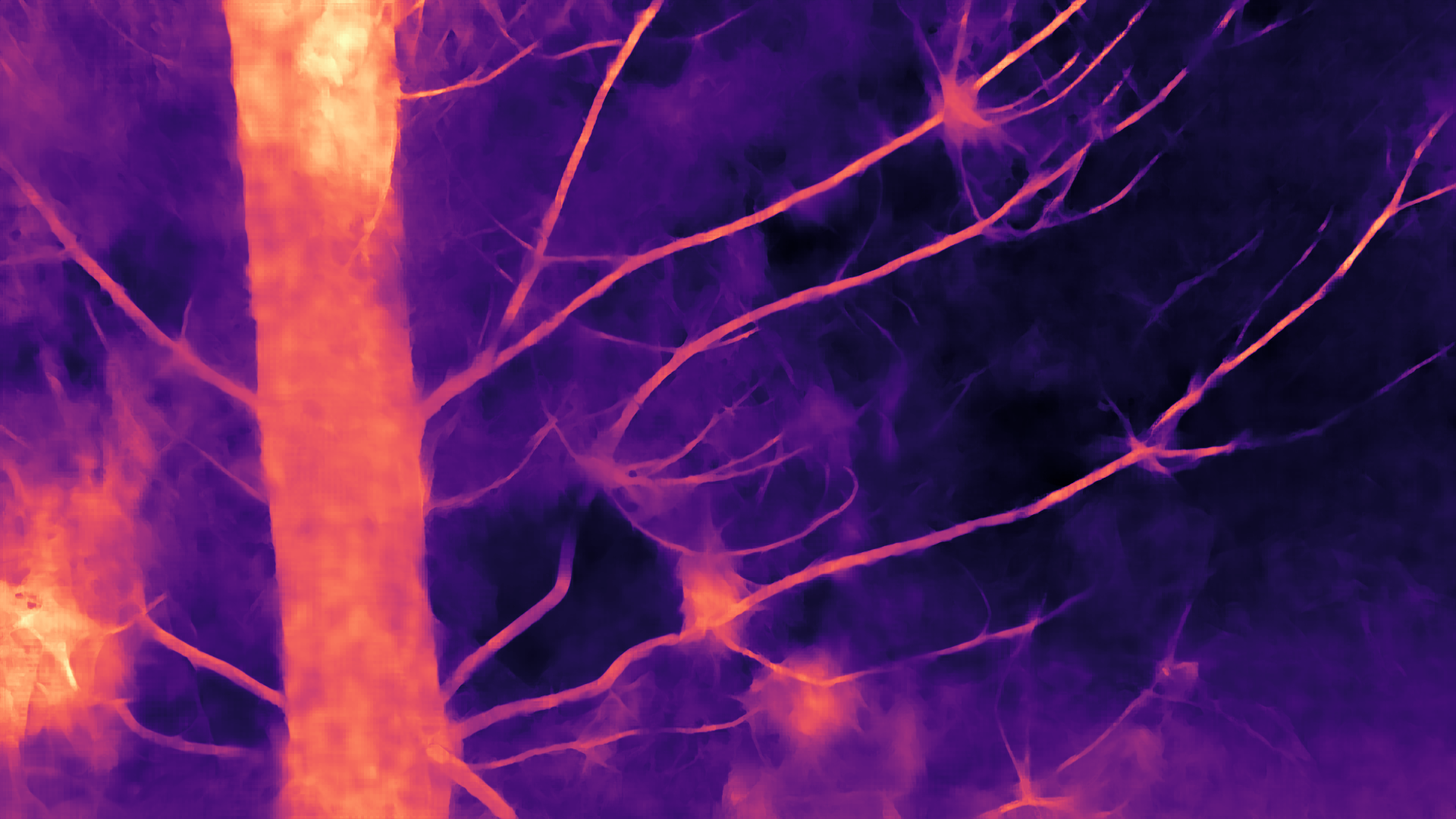}
            \caption{BANet-2D}
        \end{subfigure}
        \begin{subfigure}
            [b]{0.16\textwidth}
            \centering
            \includegraphics[width=\textwidth]{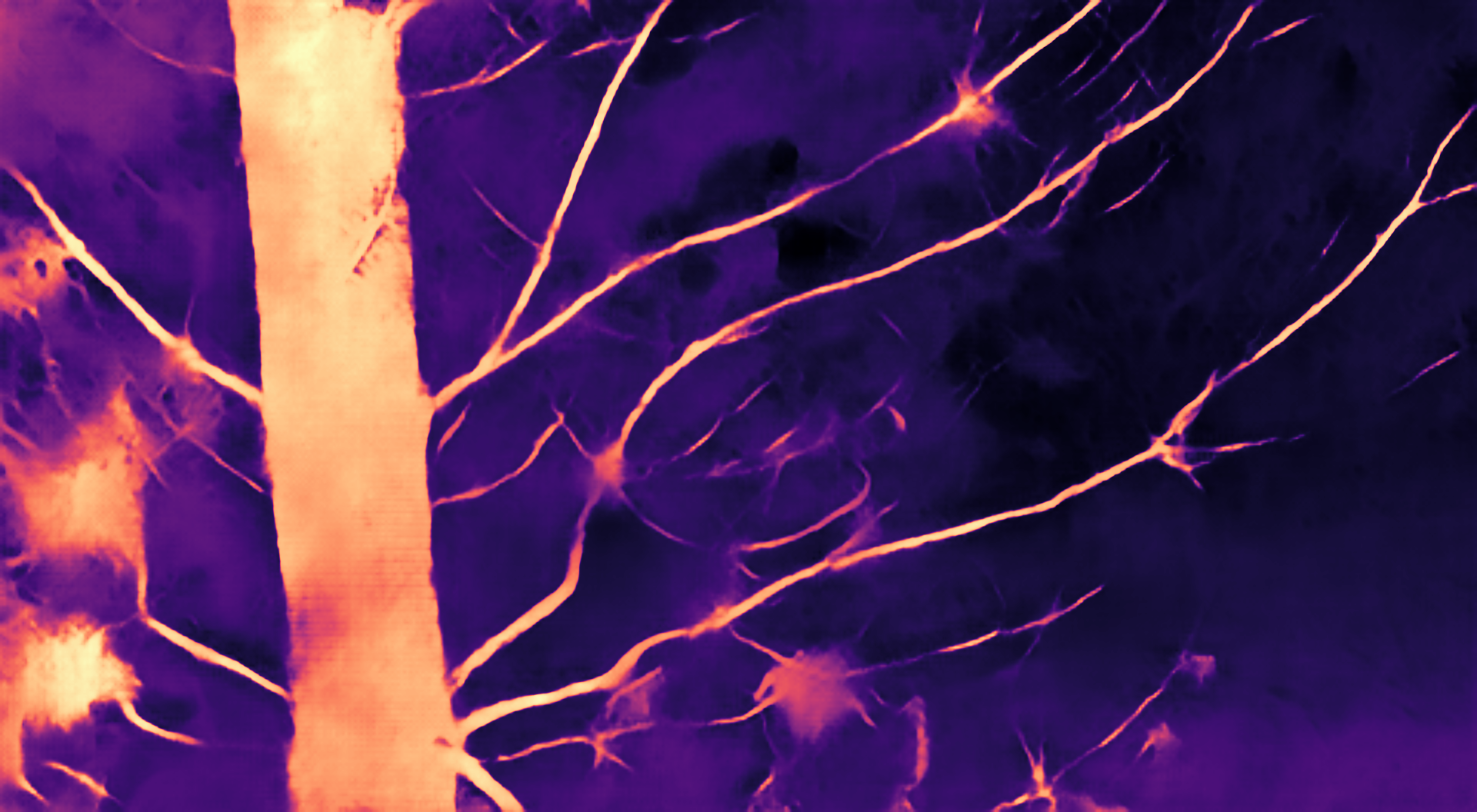}
            \caption{GwcNet}
        \end{subfigure}
        \begin{subfigure}
            [b]{0.16\textwidth}
            \centering
            \includegraphics[width=\textwidth]{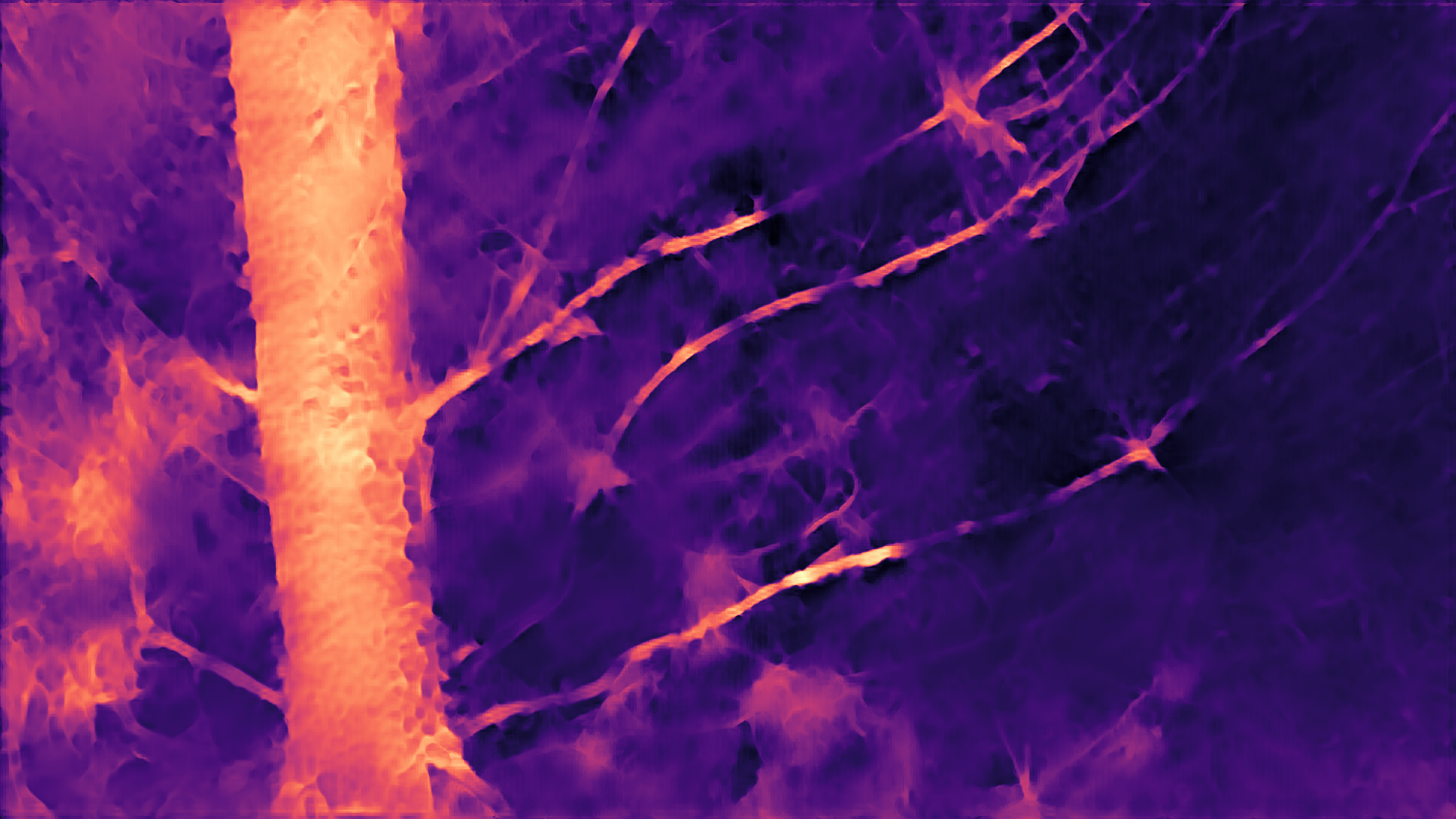}
            \caption{IGEV-RT}
        \end{subfigure}
        \begin{subfigure}
            [b]{0.16\textwidth}
            \centering
            \includegraphics[width=\textwidth]{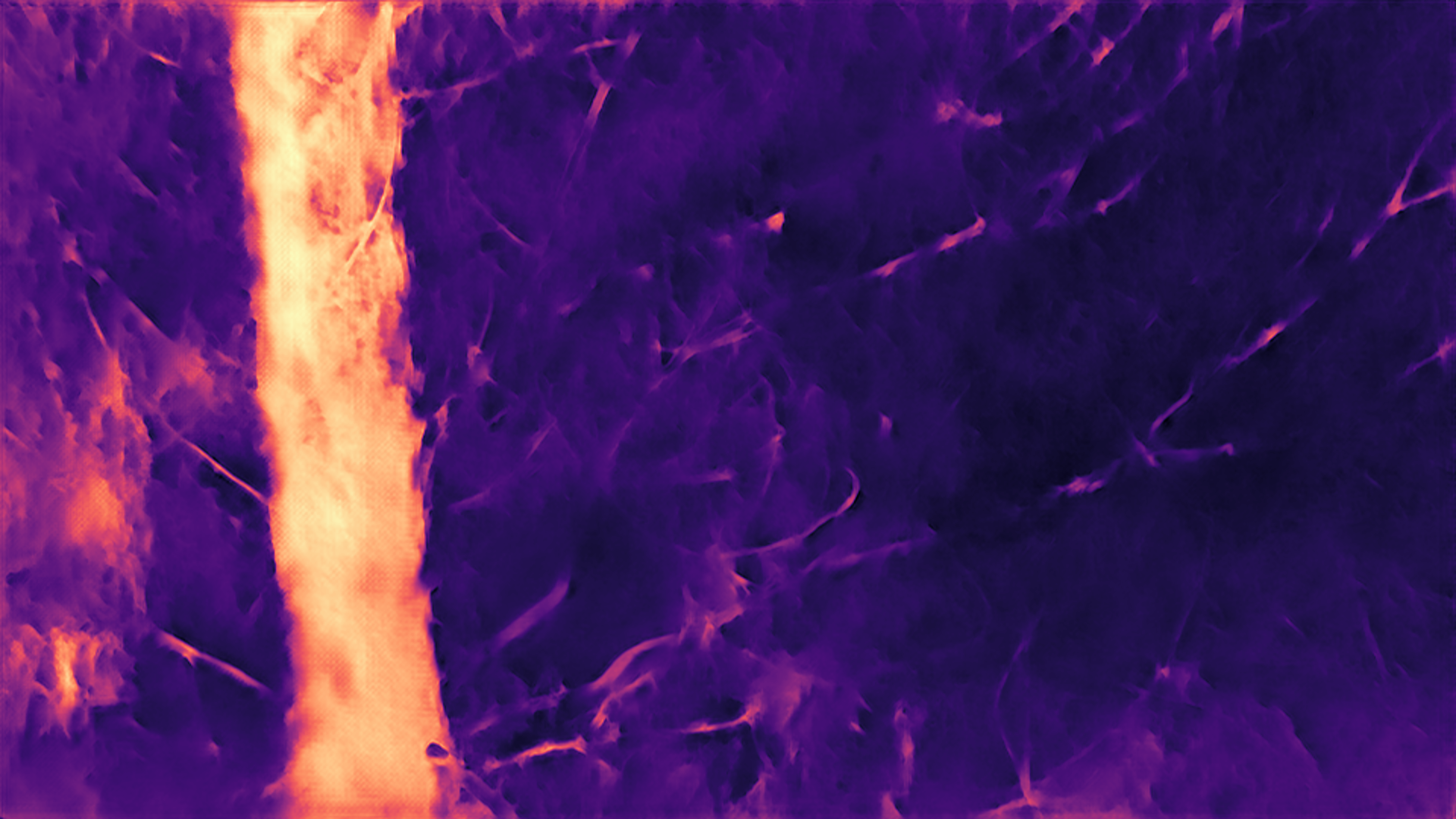}
            \caption{DeepPruner}
        \end{subfigure}
        \begin{subfigure}
            [b]{0.16\textwidth}
            \centering
            \includegraphics[width=\textwidth]{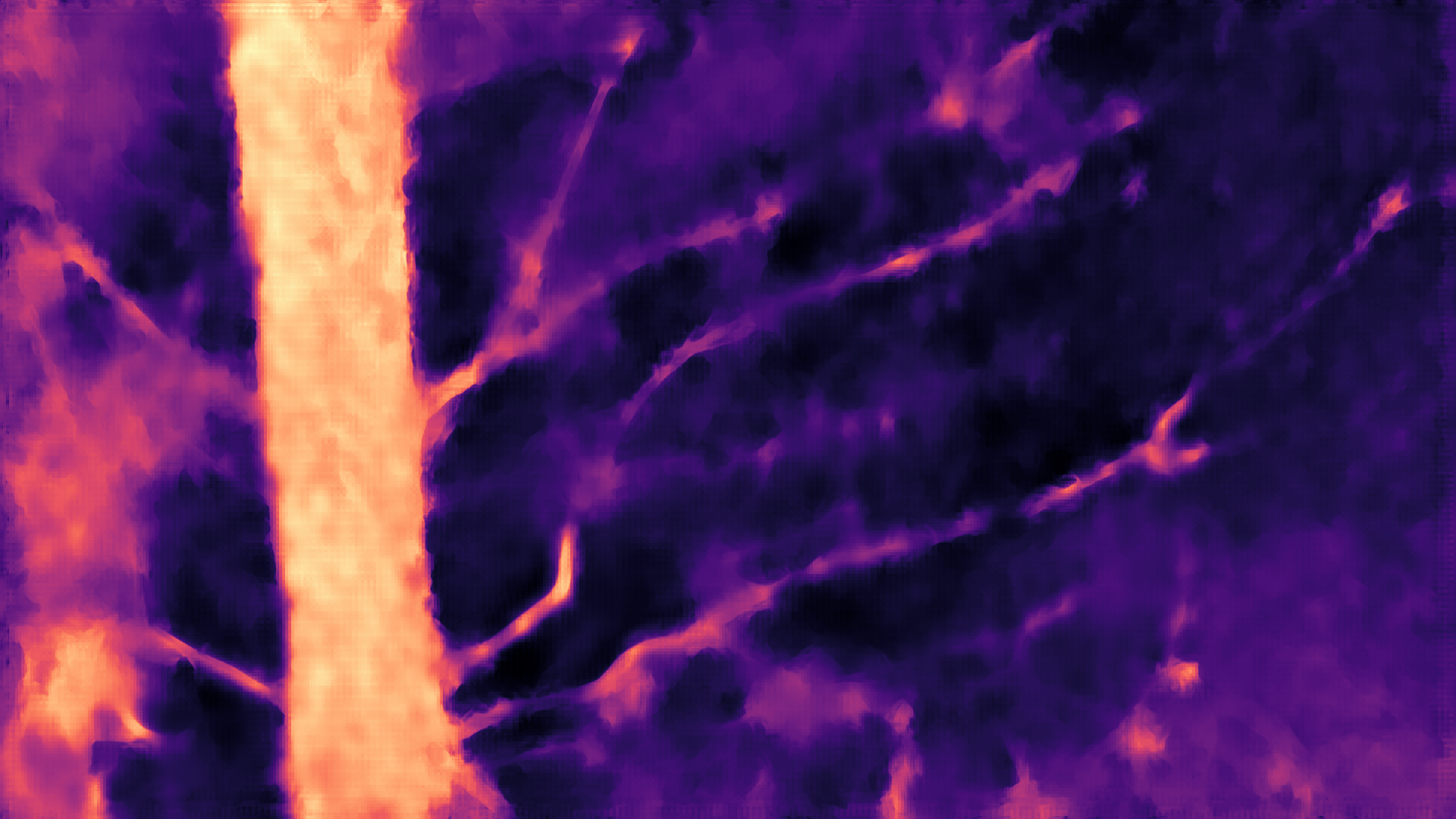}
            \caption{DCVSMNet}
        \end{subfigure}
        \begin{subfigure}
            [b]{0.16\textwidth}
            \centering
            \includegraphics[width=\textwidth]{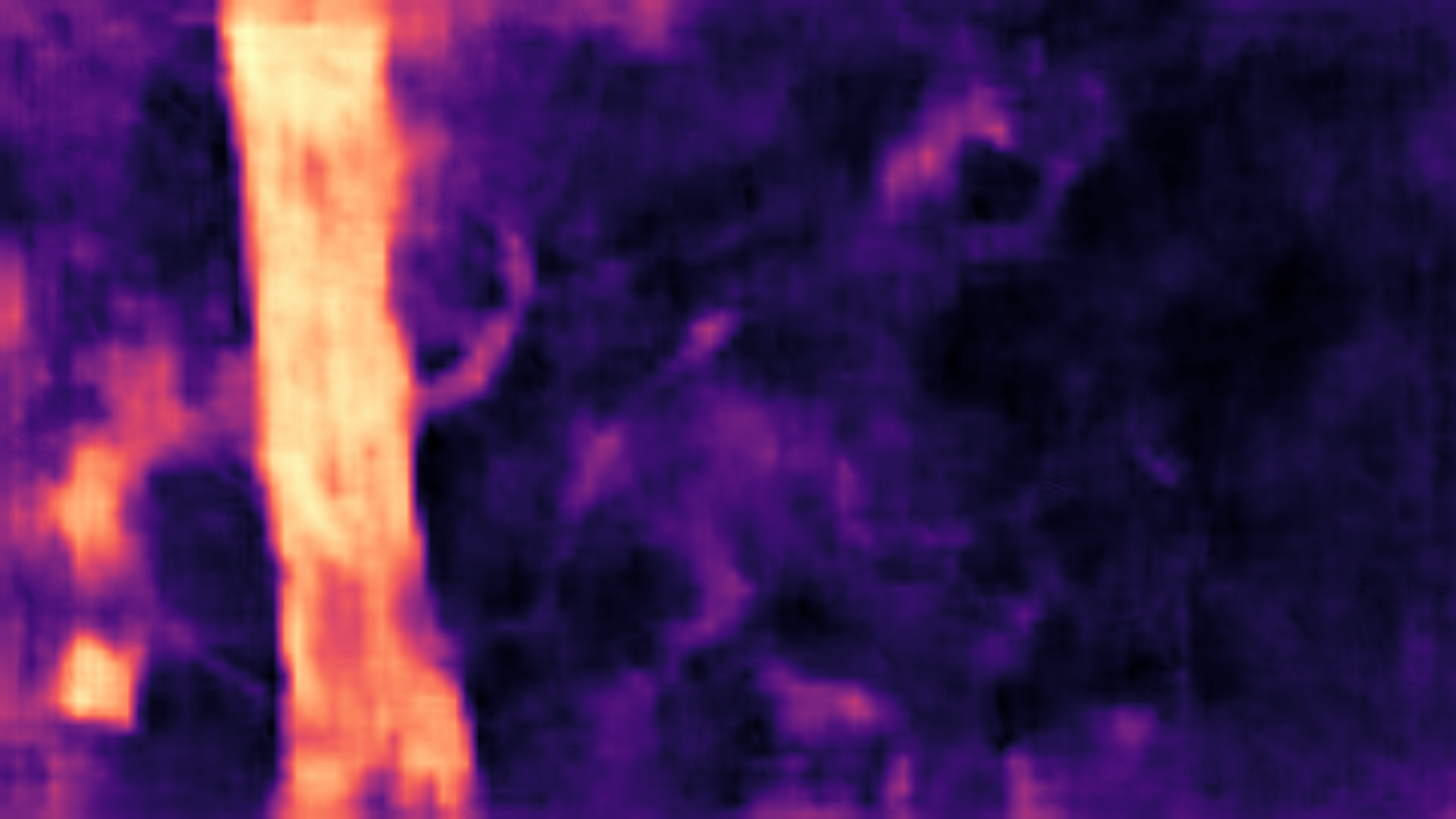}
            \caption{AnyNet}
        \end{subfigure}
        \caption{Qualitative comparison of disparity maps on a representative
        test scene. (a)~Left input image. (b)~DEFOM pseudo-ground-truth. (c)--(l)~Predictions
        from ten trained methods. BANet-3D best preserves thin branch details and
        depth boundaries, while AnyNet over-smooths fine structures. RAFT-Stereo
        captures large-scale depth layering but exhibits local artifacts.}
        \label{fig:qualitative}
    \end{figure*}

    Fig.~\ref{fig:qualitative} shows visual comparisons on test scenes with thin
    branches, blocked areas, and different lighting. Each row displays the left
    input image, the DEFOM reference disparity map, and predictions from the ten
    trained methods.

    \textbf{Branch Structure Preservation}: BANet-3D reproduces the fine branch details
    from the DEFOM reference most accurately, which aligns with its top SIFT/ORB
    match scores. RAFT-Stereo keeps the overall depth layout (matching its high
    ViTScore) but creates local distortions near thin branches. AnyNet heavily
    blurs fine structures, merging nearby branches and losing sharp depth edges.

    \textbf{Sky and Background Regions}: All trained methods produce consistent disparity
    values in the featureless sky areas. BANet-3D and MoCha-Stereo show the
    cleanest transitions between sky and tree crown edges. DeepPruner and DCVSMNet
    produce visible bleeding effects at object borders.

    \subsection{Field Deployment on UAV Platform}

    We run all ten trained models on our test drone with an NVIDIA Jetson Orin
    Super (16\,GB, its own power supply) and ZED Mini camera in real outdoor conditions.
    Unlike the offline test set, this evaluation uses live video at both 1920$\times$1080
    and 1280$\times$720 during actual flights over pine plantation canopies,
    testing how well the models handle new, unseen data in real time.

    \textbf{Power Usage}: AnyNet draws the least power ($\sim$12\,W), which helps
    extend flight time. Methods with heavy 3D processing (RAFT-Stereo, PSMNet, BANet-3D)
    use 10--20\,W more (83--167\% increase), which noticeably shortens available
    flight time---an important concern since even the Jetson's dedicated battery
    has limited capacity.

    \textbf{Heat Management}: Running RAFT-Stereo and PSMNet continuously causes
    the Jetson to overheat and slow down after about 8 minutes. AnyNet and BANet-2D
    keep steady speed throughout 30-minute flights without heat problems,
    confirming they are suitable for extended field use.
    \vspace{-1em}

    \section{Discussion}

    \subsection{Benefits of Vegetation-Specific Training}
    Our results show that training stereo networks on tree imagery with DEFOM reference
    labels produces strong, task-specific models. BANet-3D scores best across multiple
    metrics (SSIM\,=\,0.883, LPIPS\,=\,0.157, SIFT Ratio\,=\,0.274, ORB Ratio\,=\,0.162),
    showing that its edge-aware attention with 3D cost processing handles the typical
    challenges of vegetation well: thin overlapping branches, repeating leaf textures,
    and sharp depth changes.

    \textbf{Reference Label Quality}: The strong results across all ten methods
    (SSIM from 0.720 to 0.883) confirm that DEFOM predictions work well as
    training labels. Several methods reach SSIM above 0.80, which shows that
    carefully chosen model predictions~\cite{lin2026benchmark} can replace
    costly LiDAR data for training vegetation-specific stereo networks.

    \textbf{From Disparity to Depth}: Since depth is inversely related to
    disparity (Eq.~\eqref{eq:depth}), the quality of disparity maps directly affects
    depth accuracy. High SSIM and low LPIPS mean that both nearby objects (high disparity)
    and distant ones (low disparity) are reproduced well, giving reliable depth estimates
    across the full working range.

    \subsection{Architecture Insights for Forestry Applications}

    \textbf{Edge-Aware Attention Works Best}: BANet-3D's consistent lead across
    metrics suggests that edge-sensitive filtering is especially useful for
    vegetation, where depth boundaries are dense and follow complex branch shapes.
    The 3D version clearly outperforms BANet-2D (SSIM 0.883 vs.\ 0.816),
    confirming that processing the full 3D cost volume captures depth
    relationships that 2D filtering misses.

    \textbf{Rankings Depend on the Metric}: An important observation is that
    method rankings shift considerably between metrics. RAFT-Stereo gets the best
    ViTScore (0.799) but only 9th-best SSIM (0.763), while MoCha-Stereo gets 2nd-best
    SSIM (0.848) but 6th-best ViTScore (0.701). This gap shows that pixel-level similarity
    (SSIM), visual quality (LPIPS), and scene-level structure (ViTScore) measure
    quite different things---which is why we use multiple metrics in this study.

    \textbf{3D Convolution Methods}: PSMNet's strong LPIPS (0.212, 2nd best) and
    ViTScore (0.786, 3rd best) show that its multi-scale pooling captures context
    well for vegetation. GwcNet performs similarly but slightly lower, consistent
    with their relative standings on other benchmarks.

    \textbf{Best Trade-off Options}: Out of ten methods, only BANet-3D, BANet-2D,
    and AnyNet offer unbeatable quality--speed combinations. This greatly
    simplifies the choice for drone stereo systems: practitioners only need to pick
    among these three based on how much delay their application can tolerate.

    \subsection{Practical Deployment Considerations}

    \textbf{Resolution Choice}: The 720P versus 1080P speed comparison (Fig.~\ref{fig:latency_comparison})
    gives practical guidance. For closed-loop pruning control needing more than
    5\,FPS, AnyNet at 720P is essentially the only option. For pre-approach
    planning at 1--2\,FPS, BANet-2D at 720P offers much higher quality. This three-way
    trade-off between resolution, quality, and speed should guide system design.

    \textbf{Separate Power Supply}: Our Jetson Orin Super runs on its own battery,
    separate from the drone's flight battery. This is important because heavy methods
    (RAFT-Stereo: $\sim$32\,W, PSMNet: $\sim$24\,W) would otherwise cut into
    flight time significantly. Even with separate power, the dedicated battery
    still has limits, so lower-power methods (AnyNet: $\sim$12\,W) allow longer operation---a
    key factor for commercial surveys covering large plantation areas.

    \textbf{Heat Issues}: Running heavy methods at full load makes the Jetson overheat
    and slow down after 8--10 minutes. Lighter methods keep steady performance through
    entire flights, removing the need for complex heat management strategies.

    \subsection{Limitations}

    \textbf{Reference Label Ceiling}: Since our models learn from DEFOM predictions,
    they cannot be more accurate than DEFOM itself. The metrics we use (SSIM, LPIPS,
    ViTScore, SIFT/ORB) measure how closely models match DEFOM output, not absolute
    geometric accuracy. Future work should compare selected models against LiDAR
    measurements on sample scenes.

    \textbf{Limited Species and Conditions}: The Canterbury Tree Branches
    dataset covers only Radiata pine (\textit{Pinus radiata}) under New Zealand conditions.
    Whether the results hold for other tree types, climates, or seasons needs
    further testing and expanded datasets.

    \textbf{No TensorRT Optimization}: All speed measurements use standard
    PyTorch. TensorRT can speed things up by 2--5$\times$ on Jetson hardware;
    AnyNet with TensorRT could likely exceed 15\,FPS at 1080P, making true real-time
    use possible. We leave this optimization for future work.

    \section{Conclusion}

    This paper presents the first study to train and evaluate ten deep stereo matching
    networks on real tree branch images for drone-based autonomous pruning.
    Using DEFOM-generated disparity maps as training targets---where disparity
    records horizontal pixel correspondence and depth follows from $Z = fB/d$---we
    test ten methods from six design families on the Canterbury Tree Branches
    dataset. We evaluate using perceptual (SSIM, LPIPS, ViTScore) and structural
    (SIFT/ORB feature matching) metrics, with speed testing on an independently-powered
    NVIDIA Jetson Orin Super.

    Our main findings are:

    \begin{itemize}
        \item \textbf{BANet-3D gives the best quality}: It leads on four of five
            quality metrics (SSIM\,=\,0.883, LPIPS\,=\,0.157, SIFT Ratio\,=\,0.274,
            ORB Ratio\,=\,0.162). Its edge-aware attention with 3D cost processing
            fits vegetation scenes with many depth edges well.

        \item \textbf{Three methods offer unbeatable trade-offs}: Only BANet-3D (best
            quality), BANet-2D (balanced), and AnyNet (fastest) sit on the best trade-off
            boundary; all other methods are outperformed.

        \item \textbf{Resolution affects speed}: Comparing 720P and 1080P processing
            on the Jetson Orin Super provides practical guidance for choosing the
            right resolution--quality--speed balance for different pruning tasks.

        \item \textbf{Confirmed in the field}: Testing on a live drone with independently-powered
            Jetson and ZED Mini shows that AnyNet and BANet-2D keep steady performance
            through full-length flights without overheating.
    \end{itemize}

    We plan to publicly release the Canterbury Tree Branches dataset with DEFOM reference
    labels and trained model weights to support further research in drone-based forestry.
    Future work will explore TensorRT optimization for faster processing, self-supervised
    learning techniques, frame-to-frame consistency for video input, and combining
    these models with branch detection and segmentation systems~\cite{lin2024branch,lin2025segmentation}
    to build complete autonomous pruning systems.

    \section*{Acknowledgments}
    This research was supported by the Royal Society of New Zealand Marsden Fund
    and the Ministry of Business, Innovation and Employment. We thank the forestry
    research stations for data collection access.

    \bibliographystyle{IEEEtran}
    
\end{document}